\documentclass{article}
\usepackage[preprint,nonatbib]{neurips_2021}

\usepackage[utf8]{inputenc} 
\usepackage[T1]{fontenc}    
\usepackage{hyperref}       
\usepackage{url}            
\usepackage{booktabs}       
\usepackage{amsfonts}       
\usepackage{nicefrac}       
\usepackage{microtype}      
\usepackage{xcolor}         

\usepackage{times}
\usepackage{epsfig}
\usepackage{graphicx}
\usepackage{amsmath}
\usepackage{amssymb}

\usepackage{multirow}
\definecolor{Rosolic}{cmyk}{0.00,1.00,0.50,0}
\definecolor{bleudefrance}{rgb}{0.19, 0.55, 0.91}
\definecolor{green}{cmyk}{0.3,0.2,0.95,0.0}
\definecolor{brown}{cmyk}{0.4,0.7,1.0,0.5}
\definecolor{dblue}{cmyk}{1,0.97,0.35,0.0}
\definecolor{Cyan}{cmyk}{1,0,0,0}
\definecolor{Azure}{rgb}{0.0, 0.5, 1.0}

\newcommand{\D}{\mathcal{D}}
\newcommand{\loss}{\mathcal{L}}
\newcommand{\R}{\mathbb{R}}

\newcommand{\attn}{\mathrm{ATTN}}
\newcommand{\ffn}{\mathrm{FFN}}

\title{ADeLA: Automatic Dense Labeling with Attention for \\
Viewpoint Adaptation in Semantic Segmentation}

\author{%
  Yanchao Yang\thanks{Equal contributions.} \ \thanks{Corresponding author.} \\
  Stanford University\\
  \And
  Hanxiang Ren\footnotemark[1] \\
  Zhejiang University \\
  \And
  He Wang \\
  Stanford University \\
  \And
  Bokui Shen \\
  Stanford University \\
  \And
  Qingnan Fan \\
  Stanford University \\
  \And
  Youyi Zheng\footnotemark[2] \\
  Zhejiang University \\
  \And
  C. Karen Liu \\
  Stanford University \\
  \And
  Leonidas Guibas \\
  Stanford University
}

\begin{document}

\maketitle
\vspace{-0.4cm}

\begin{figure}[!h]
   \centering
   \includegraphics[width=0.93\linewidth]{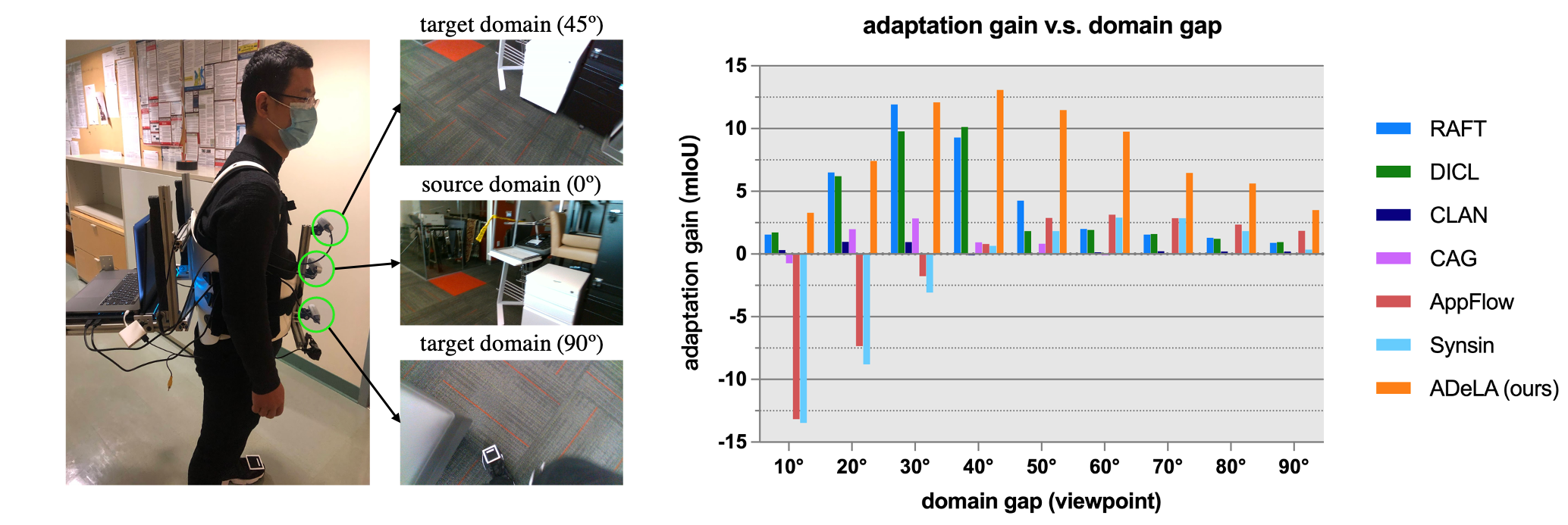}
   \vspace{-0.2cm}
\caption{Left: An assistive exoskeleton with multiple cameras towards different viewpoints can be used to improve the user mobility. However, the performance of the semantic segmentation network trained on the forward viewpoint (typical view of existing datasets) drops sharply when tested on different viewpoints (Tab.~\ref{tab:benchmarking}). Right: Adaptation gain obtained by state-of-the-art methods for adapting between viewpoints. Our method consistently achieves positive adaptation gain and works robustly towards substantial viewpoint change, \textit{e.g.}, perpendicular viewing angles.}
\label{fig:teaser}
\end{figure}

\begin{abstract}\label{sec:abstract}
We describe an unsupervised domain adaptation method for image content shift caused by viewpoint changes for a semantic segmentation task. Most existing methods perform domain alignment in a shared space and assume that the mapping from the aligned space to the output is transferable. However, the novel content induced by viewpoint changes may nullify such a space for effective alignments, thus resulting in negative adaptation. Our method works without aligning any statistics of the images between the two domains. Instead, it utilizes a view transformation network trained only on color images to hallucinate the semantic images for the target. Despite the lack of supervision, the view transformation network can still generalize to semantic images thanks to the inductive bias introduced by the attention mechanism. Furthermore, to resolve ambiguities in converting the semantic images to semantic labels, we treat the view transformation network as a functional representation of an unknown mapping implied by the color images and propose functional label hallucination to generate pseudo-labels in the target domain. Our method surpasses baselines built on state-of-the-art correspondence estimation and view synthesis methods. Moreover, it outperforms the state-of-the-art unsupervised domain adaptation methods that utilize self-training and adversarial domain alignment. Our code and dataset will be made publicly available.
\end{abstract}

\begin{figure}[!t]
\begin{center}
   \includegraphics[width=0.92\linewidth]{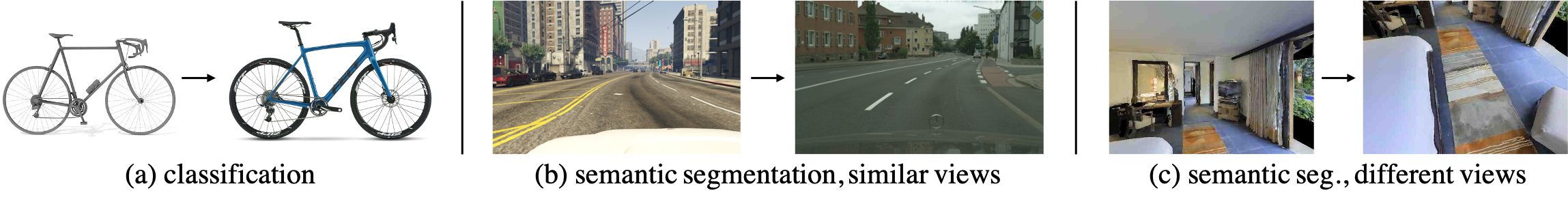}
\end{center}
\vspace{-0.35cm}
\caption{(a): image classification where image style and viewpoint are nuisances \cite{peng2018syn2real,saenko2010adapting}. (b): semantic segmentation at similar views where image style is the major nuisance for domain gaps \cite{peng2017visda,hoffman2018cycada}. (c): semantic segmentation with changing view (\textit{e.g.}, forward to downward), a nuisance that should not be aligned away. We focus on UDA for viewpoint change in semantic segmentation.}
\vspace{-0.1cm}
\label{fig:different-tasks}
\end{figure}

\section{Introduction}\label{sec:intro}
\vspace{-0.3cm}

Parsing the environment from multiple viewing angles to arrive at a comprehensive understanding of the surroundings is critical in many settings for autonomous agents or assistive robots. An example is the case of a human exoskeleton instrumented with multiple cameras as shown in Fig.~\ref{fig:teaser}, where training a scene parsing network that performs well at multiple viewpoints is key to estimating the walkable floor surface and preventing falls and other accidents. 
However, viewpoint changes across cameras induce significant domain gaps -- as a result, a scene parsing network trained with annotations in only one view usually encounters a large performance drop on another (Tab.~\ref{tab:benchmarking}).
We aim at reducing this domain gap by adapting the network from the view with rich annotations (source) to views where no annotation is available (target), i.e., unsupervised domain adaptation (UDA).

Most UDA methods build on the idea that an alignment in a shared latent space helps the task-specific network trained in the source domain generalize to the target. Despite its effectiveness in practice, domain alignment generally assumes (sufficient) invariance exists for the task, which can be computed through the alignment, so that the mapping from the aligned space to the output is transferable across domains. 
For example, if one converts the images in Fig.~\ref{fig:different-tasks} (a) and (b) into edge maps, the discrepancy between domains may diminish, so the task networks trained on the source edge maps would also work on the target.
However, the domain discrepancy we are considering here is mainly the content shift caused by the viewpoint change. As dense scene parsing (semantic segmentation) is viewpoint elevation-dependent, any alignment that learns away viewpoint will result in (insufficient) invariances which are not adequate or suitable for the task, thus inducing negative adaptation (Fig.~\ref{fig:teaser}).

We break this conundrum by hallucinating the target semantic images using their source counterparts. Instead of aligning features between domains, our method employs a view transformation network that outputs the target semantic image, conditioned on a source semantic image and a pair of regular color images. The hallucinated semantic images are then converted to semantic labels to adapt the task network. Since semantic images in the target domain are missing, the only supervision for the view transformation network is the color images. One can train a network to hallucinate using the color images and apply it to the semantic images. However, without a proper inductive bias, the view transformation network would completely fail on semantic images due to their different structures. We propose that the right inductive bias is to encourage learning \emph{spatial transportation} instead of transformation in color space. Further, we introduce a novel architecture for view transformation where the desired inductive bias is injected via an attention mechanism. 
To combat noise in the hallucination and better decode the semantic labels, we treat the view transformation network as a functional representation of an unknown mapping signified by the color images. Accordingly, we propose a functional label hallucination strategy that generates the soft target labels by taking in the indicator functions of each class. The proposed decoding strategy improves the label accuracy by a large margin and makes the labels more suitable for adaptation by incorporating uncertainties.

Due to the lack of datasets in semantic segmentation whose domain gaps are mainly from viewpoint change, we also propose a new dataset where the viewpoint is varied to simulate different levels of content shift. We perform an extensive study of the state-of-the-art UDA methods, and show that the adaptation gain of domain alignment vanishes quickly when the viewpoint-induced domain gap increases. Moreover, we show that the soft labels from our method are superior to those from the state-of-the-art dense correspondence estimation and view synthesis methods, even if the ground-truth camera poses are made available to them. Our method consistently achieves the best adaptation gains across different target domains, even for perpendicular viewing angles, demonstrating the effectiveness of the proposed architecture (inductive bias) and functional transportation strategy.

\vspace{-0.1cm}
\section{Related work}
\vspace{-0.1cm}

We focus on unsupervised domain adaptation (UDA) methods for the pixel-level prediction task of semantic segmentation.
The core ingredient of unsupervised domain adaptation is to reduce the domain shift between the source and the target data distributions \cite{patel2015visual,csurka2017domain,wang2018deep,glorot2011domain,fernando2013unsupervised,baktashmotlagh2013unsupervised}, 
where the domain shift can be measured by maximum mean discrepancy \cite{geng2011daml,long2015learning} or central moment discrepancy \cite{zellinger2017central}. 
Deep learning based methods resort to adversarial measurements,
where discriminator networks are used to confuse the two domains \cite{sener2016learning,tzeng2017adversarial,motiian2017unified,saito2017asymmetric,shu2018dirt,kumar2018co} in a shared feature space.
In contrast to classification, feature space alignment is much less effective for pixel-level prediction tasks like semantic segmentation \cite{luo2019significance,sankaranarayanan2018learning}, due to the difficulties in keeping the aligned features informative about the spatial structure of the output.

The recent success of unsupervised domain adaptation for semantic segmentation mainly relies on image-to-image translation \cite{zhu2017unpaired,liu2017unsupervised,yi2017dualgan} where the goal is to reduce the style difference between domains while preserving the underlying semantics \cite{zhang2017curriculum,hoffman2018cycada,li2019bidirectional}.
Multi-level feature alignment is proposed in \cite{wu2018dcan} and \cite{gong2019dlow} introduces intermediate styles that gradually close the gap.
A disentanglement of texture and structure is also beneficial \cite{chang2019all},
and \cite{kim2020learning} performs style randomization to learn style invariance.
To ease the difficulty in adversarial training, \cite{yang2020fda} proposes a style transfer via Fourier Transform, which enforces semantic consistency, and \cite{yang2020phase} directly regularizes the image translation module using a phase preserving constraint.
On the other hand, \cite{zhang2019category,du2019ssf,luo2019taking,wang2020differential,dong2020cscl} propose class-wise alignments, given that each of the semantic classes may possess a different domain gap.
Similarly, \cite{tsai2019domain} proposes patch-wise alignment, and \cite{huang2020contextual} utilizes local contextual-relations for a consistent adaptation.
The alignment can also be performed in the output space 
\cite{vu2019advent}, or in a curriculum manner. For example, \cite{pan2020unsupervised} employs inter and intra domain adaptation with an easy-to-hard split, and \cite{li2020content} pre-selects source images that share similar content with the target.
With aligned domains, self-training using pseudo labels can be utilized to further close the gap \cite{zhang2021prototypical,li2019bidirectional,yang2020fda}.

Our method tackles the domain gap caused by different camera views, which renders the image space alignment ineffective as the domain gap is mainly content shift but not the style difference. 
Unlike cross-view image classification \cite{rahmani2017learning,zhang2017view,deng2018image,bak2018domain,fu2019self}, aligning domains of different viewpoints for pixel-level prediction tasks is ill-posed, since the task is indeed view dependent \cite{combes2020domain}.
The most relevant are \cite{di2020sceneadapt,coors2019nova}, which again resort to adversarial domain alignment. Additionally, \cite{coors2019nova} requires known camera intrinsics and extrinsics.
Note, both assume the viewpoint change is small or there is a large overlap between the two views,
therefore the applicability to a broader setting is limited, whereas our method is not constrained by any of these assumptions. 
Also related is novel view synthesis \cite{zhou2018stereo,sitzmann2019scene,flynn2019deepview,choi2019extreme}, particularly, single view synthesis \cite{zhou2016view,tucker2020single,wiles2020synsin}, where multiple posed images of the same scene are needed during training. Hence, if the goal is to synthesize semantic images of a different view, the target domain's semantic images are needed, which, however, are not available in our problem setting.
Another related topic is dense correspondence estimation \cite{teed2020raft,wang2020displacement,zhao2020maskflownet}, which can be used to warp labels to help adaptation between domains.

\begin{figure}[!t]
   \centering
   \includegraphics[width=0.9\linewidth]{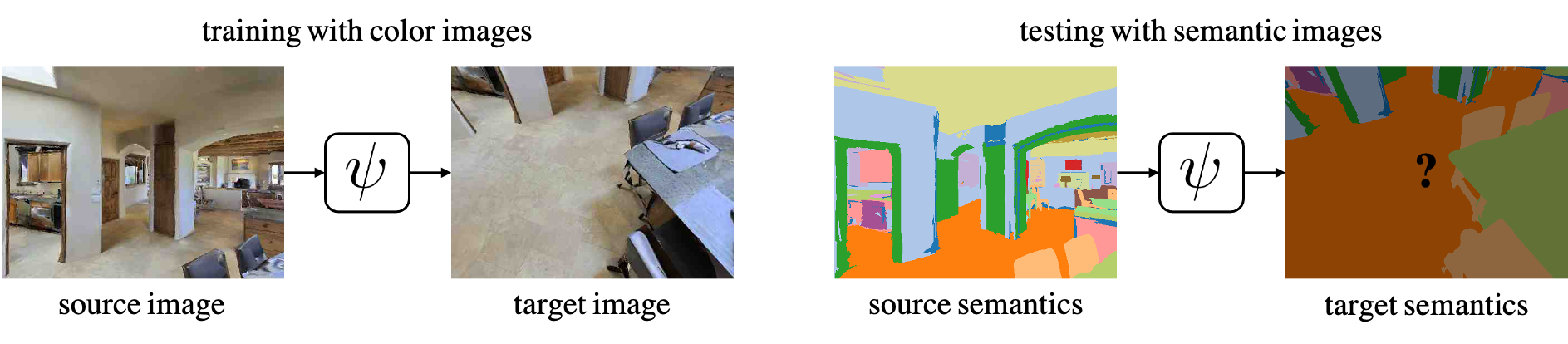}
   \vspace{-0.3cm}
\caption{Left: a network $\psi$ is trained to hallucinate color images from the source to the target and is never exposed to semantic images; Right: $\psi$ is directly applied on the corresponding source semantic image to hallucinate the target semantic image to provide training labels for the target domain.}
\vspace{-0.2cm}
\label{fig:motivation}
\end{figure}

\vspace{-0.1cm}
\section{Method}
\vspace{-0.1cm}

Let $\D^s=\{(x^s_i,y^s_i)\}_{i=1}^n$ be the source dataset collected at the source viewpoint $s$, 
where $x^s_i\in\R^{h\times w\times 3}$ is an RGB image, 
and $y^s_i\in\R^{h\times w\times 3}$ is the corresponding semantic image that is color coded according to the semantic labels for visualization\footnote{One can always convert the semantic image to integer semantic labels using nearest neighbour search.}.
Further, let $\D^\tau=\{x^\tau_i\}_{i=1}^n$ be the target dataset collected at the target viewpoint $\tau$, whose semantic label/image $y^\tau_i$ is to be predicted.
In order to make our method generally applicable,
we assume no knowledge about the viewpoints $s,\tau$ except that $x^s_i\in\D^s$ and $x^\tau_i\in\D^\tau$ are synchronized. 
Therefore, the domain gap between $\D^s$ and $\D^\tau$ comes from the viewpoint difference in our problem setting.
Moreover, the synchronized source and target view images may or may not share co-visible regions, which is determined by the difference between the two views.
Please see Fig.~\ref{fig:datasets} for examples of the source domain and target domains with increasing viewpoint changes.

Similar to unsupervised domain adaptation, our ultimate goal is to train a semantic segmentation network $\phi:\mathbf{x} \rightarrow \mathbf{y}$ given only the annotations from the source dataset $\D^s$ so that $\phi$ can perform well on the target dataset $\D^\tau$ with the presence of domain gaps. 
The domain gap we are considering here is mainly the content shift induced by different viewing angles, \textit{i.e.}, the discrepancy in the output structures, which violates the assumptions made by most unsupervised domain adaptation methods that rely on either image space alignment or feature space alignment, or both \cite{zhang2021prototypical,luo2019taking,zhang2019category,yang2020fda,kim2020learning,li2020content,vu2019advent}.
Instead of aligning distributions of any kind between the two domains, which may result in negative adaptation gains (Fig.~\ref{fig:teaser}, right),
we resort to a network that can hallucinate the target view semantic images ($y^\tau$) from the source ($y^s$) guided by the color images ($x^s,x^\tau$).
Specifically, we want to have a network $\psi: \mathbf{y} \times \mathbf{x} \times \mathbf{x} \rightarrow \mathbf{y}$, whose output $\psi(y^s_i; x^s_i, x^\tau_i)$ can be used as pseudo ground-truth for improving $\phi$ to make better predictions on $\D^\tau$.

\begin{figure}[!t]
   \centering
   \includegraphics[width=0.8\linewidth]{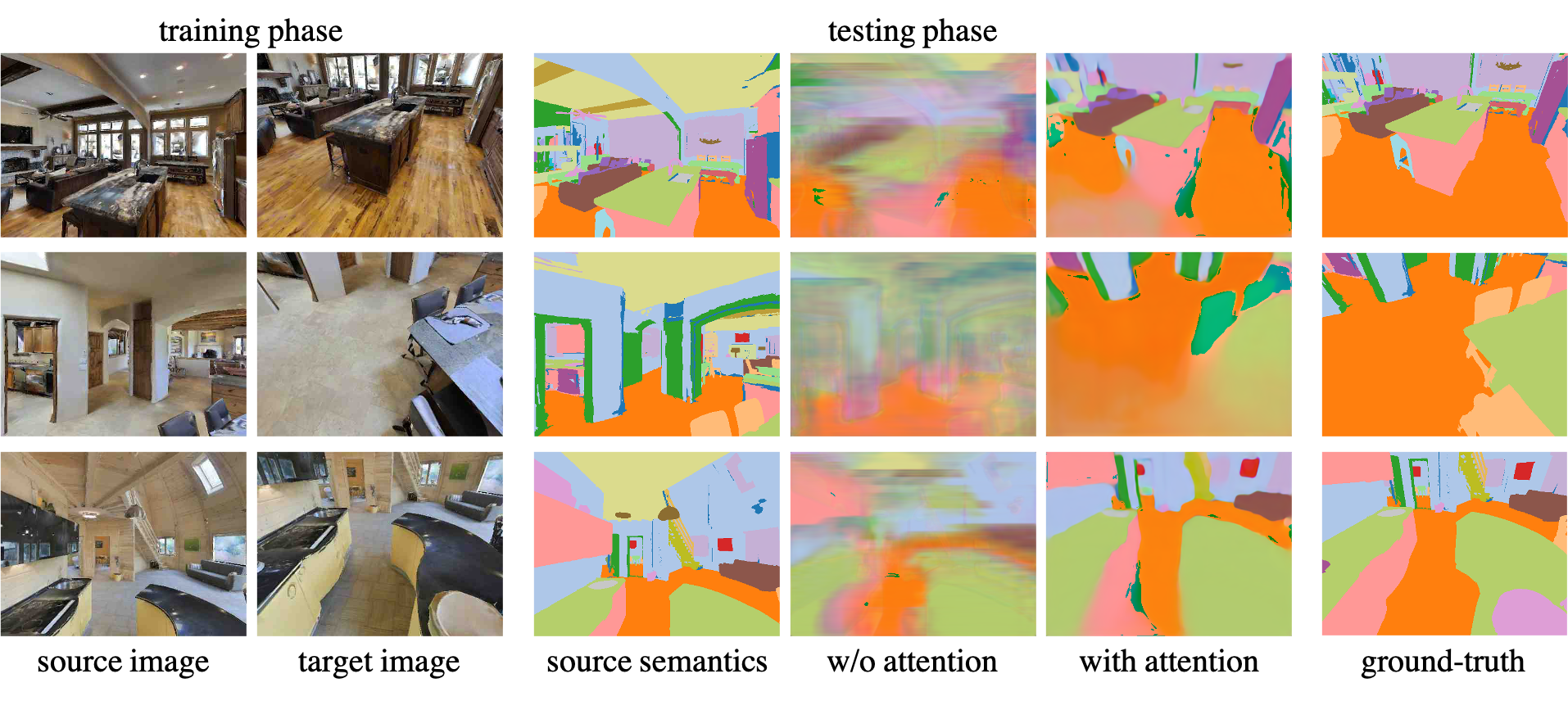}
   \vspace{-0.1cm}
\caption{Left to right: source and target color images (1st,2nd) for training the view transformation. During testing, source semantic images (3rd) are inputs to the network. The hallucinated semantic images by the network without attention (4th) are inaccurate and not consistent with the target images; however, the hallucinations from the network with attention (5th) are sharp and more precise.}
\vspace{-0.2cm}
\label{fig:attention-is-effective}
\end{figure}

\vspace{-0.1cm}
\subsection{Auto-labeling with attention}
\vspace{-0.1cm}
\label{sec:inductive-bias-attn}

Looking at a pair of color images $x^s_i, x^\tau_i$ shown in Fig.~\ref{fig:motivation}, one could hallucinate to some extent the target semantic image $y^\tau_i$ associated with $x^\tau_i$ given the source semantic image $y^s_i$.
On the other hand, if a network learns how to hallucinate the target image $x^\tau_i$ from the source image $x^s_i$, we would expect it to make a reasonable hallucination of the target semantic image $y^\tau_i$ from the source semantic image $y^s_i$, since $x^s_i$ and $y^s_i$ are simply two different appearances of the same geometry.
However, without a proper inductive bias, a network trained to hallucinate color images between different views may fail completely when tested on semantic images due to their statistical difference.

To validate, we train a UNet \cite{ronneberger2015u} $\psi^{unet}$, which is widely used for image transformation and dense prediction, to hallucinate $x^\tau_i$ from $x^s_i$, with $\tau$ fixed and L1 loss as the training objective, \textit{i.e.}, $\hat{x}^\tau_i = \psi^{unet}(x^s_i)$. 
After training, we directly test $\psi^{unet}$ on the semantic images to check if $\psi^{unet}(y^s_i)$ is similar to $y^\tau_i$.
As shown in Fig.~\ref{fig:attention-is-effective}, the UNet trained on color images for view transformation does not generalize well to semantic images, which confirms the difficulties to perform novel appearance hallucination of a seen view, even if the geometry is unchanged.

We propose that the key to generalizing to novel appearance is to bias the network towards learning spatial transportation instead of color transformation.
For example, the network needs incentives to learn where the color should be copied to in the target view instead of how the color should change to form the target view.
If so, the view hallucination should generalize to any novel appearance since the color transformation may depend on domains while the spatial transportation conditioned on the same views and geometry is invariant.

{\bf Biasing towards transportation with attention.} 
The self-attention mechanism proposed in \cite{vaswani2017attention} represents a layer that processes the input by first predicting a set of keys ($K$) and a set of queries ($Q$), whose dot-products are then used to update a set of values ($V$) to get the output (updated values):
\begin{equation*}
    \attn(Q,K,V) = softmax(\dfrac{QK^T}{\sqrt{d_k}})V
\end{equation*}
By examining how a single output value $v'_i$ is computed, we can see why attention helps to bias towards spatial transportation that facilitates the generalization of the hallucination.
Let $q_i$ be the corresponding query for $v'_i$, and $[k_1,k_2,...,k_m]$ be the keys, then $v'_i = \sum_{j=1}^m \alpha_j\cdot v_j$, with $\alpha_j$'s the elements of $softmax([k_1 q_i^T,k_2 q_i^T,...,k_m q_i^T])$ (scaling factor omitted for simplicity).
Note if $q_i$ is extremely similar to a certain key, \textit{e.g.}, $k_{j^*}$, but dis-similar to the other keys, we may write $v'_i \approx v_{j^*}$. This signals that the attention is transporting values from different locations to $i$ through the weighted summation. In the extreme case, it can even stimulate point-wise transportation of the values.

To verify the effectiveness of attention in hallucinating labels (novel appearance), we simply reorganize the tunable parameters in the UNet $\psi^{unet}$ such that the convolutional layers near the bottleneck are now replaced by attention layers of the same capacity. 
We term it as $\psi^{attn}$ and train it to hallucinate the target color images from the source color images, \textit{i.e.}, $\hat{x}^\tau_i = \psi^{attn}(x^s_i)$, 
and test it on the semantic images.
As shown in Fig.~\ref{fig:attention-is-effective} (5th column),
$\psi^{attn}$ can hallucinate reasonable target semantic images even it is only trained on color images. 
Given the effectiveness of the inductive bias introduced by the attention mechanism in label hallucination for a single target view,
we now detail our view hallucination network for multiple target views and the technique that we propose to generate soft labels for adaptation to different target domains.

\begin{figure}[!t]
   \centering
   \includegraphics[width=0.88\linewidth]{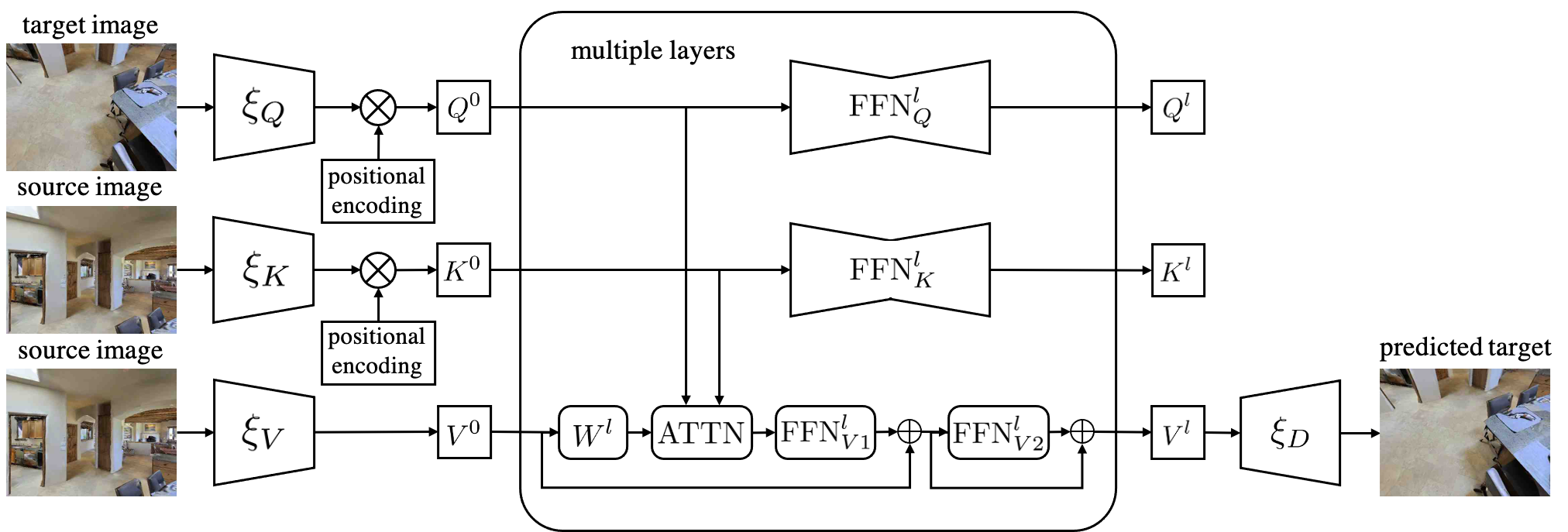}
   \vspace{-0.1cm}
\caption{The proposed architecture for hallucinating arbitrary target domains, which is specified by the input target image (top left). During testing, the input to the value encoder $\xi_V$ (bottom left) is replaced by the source semantic image. Please refer to Sec.~\ref{sec:multi-tar} for detailed explanations.}
\vspace{-0.3cm}
\label{fig:net-mutiple}
\end{figure}

\vspace{-0.1cm}
\subsection{Labeling multiple target domains} \label{sec:multi-tar}
\vspace{-0.2cm}

Here we specify the proposed network architecture that can seamlessly work for different target views, \textit{e.g.}, the target domain is a mixture of views, which eliminates the need to train separate networks.
Again, the view hallucination network $\psi(x_V; x_K,x_Q)$ takes in a pair of color images, which guide $\psi$ to predict the target view from the source whose appearance is determined by either the source color image or the source semantic image, \textit{i.e.}, $\hat{x}^\tau_i = \psi(x^s_i; x^s_i,x^\tau_i)$ during training or $\hat{y}^\tau_i = \psi(y^s_i; x^s_i,x^\tau_i)$ during testing.
As illustrated in Fig.~\ref{fig:net-mutiple}, we let $x_Q = x^\tau_i$, $x_K = x^s_i$ and $x_V = x^s_i$, which are lifted to query, key and value features through the following procedure:
\begin{align*}
    Q^0 &= \xi_Q(x_Q) [\mathbf{1}; u_{pos}; v_{pos}]\\
    K^0 &= \xi_K(x_K) [\mathbf{1}; u_{pos}; v_{pos}]\\
    V^0 &= \xi_V(x_V)
\end{align*}
here $\xi_Q, \xi_K, \xi_V$ are separate encoders with strided convolutions to reduce the spatial dimensions of the features, and $u_{pos}, v_{pos}$ are fixed positional encodings that represent the normalized image grids (horizontal and vertical).
These lifted features are followed by $L$ layers described below:
\begin{align}
    Q^l &= \ffn_Q^l(Q^{l-1})\\
    K^l &= \ffn_K^l(K^{l-1})\\
    \hat{V}^l &= \attn(Q^{l-1},K^{l-1},V^{l-1}W^{l})\\
    \bar{V}^l &= \ffn_{V1}^l(\hat{V}^l) + V^{l-1}\\
    V^l &= \ffn_{V2}^l(\bar{V}^{l}) + \bar{V}^{l}\\
    x_Q^{l} &= \xi_D(V^l)
\end{align}
where $\ffn_Q^l,\ffn_K^l$ are two feed-forward networks of downsampling and upsampling convolutional layers with layernorm to maintain the size of the updated keys and queries. 
And the feed-forward networks $\ffn_{V1}^l,\ffn_{V2}^l$ are simply convolutional layers whose stride is equal to one.
Note, for each $V^l$, we apply the shared decoder $\xi_D$ to map it to the image space, and $x_Q^L$ will be the final output of the proposed view transformation network $\psi$.

{\bf Training loss and data augmentation.} 
For training the network $\psi(x_V; x_K,x_Q)$ in Fig.~\ref{fig:net-mutiple}, we apply color jittering to the input images.
Specifically, the hue of $x^s_i, x^\tau_i$ are perturbed by a random factor to get $x_Q$ and $x_K$, and by a different factor to get $\bar{x}_Q$ and $x_V$, where $\bar{x}_Q$ is the expected output of $\psi(x_V; x_K,x_Q)$.
Different hue perturbations can help prevent information leakage, since now $x_Q$ (input) and $\bar{x}_Q$ (expected output) are not identical, yet the consistency between $x_V$ and $\bar{x}_Q$ is preserved to enable meaningful hallucination.
In addition, we apply the same color permutation to $x_V$ and $\bar{x}_Q$, \textit{e.g.}, red to green and yellow to blue, to further prevent information leakage from $x_Q$ to the output.
More details can be found in the appendix.
The training loss for $\psi$ is:
\begin{equation}
    \loss^{\psi} = \sum_{x_Q\in\{\D^\tau\}}\sum_{l=1}^L \lambda_l \| x_Q^l - \bar{x}_Q \|_1
\end{equation}
with $\lambda_l$ the weighting coefficient for the $l$-th layer's output $x_Q^l$, which is decoded from $V^l$, and we set $\lambda_l = 2^{-(L-l)}$ so that early predictions are weighted less. Note, $x_K,x_V$ can be indexed by $x_Q$.

\begin{figure}[!t]
   \centering
   \includegraphics[width=0.81\linewidth]{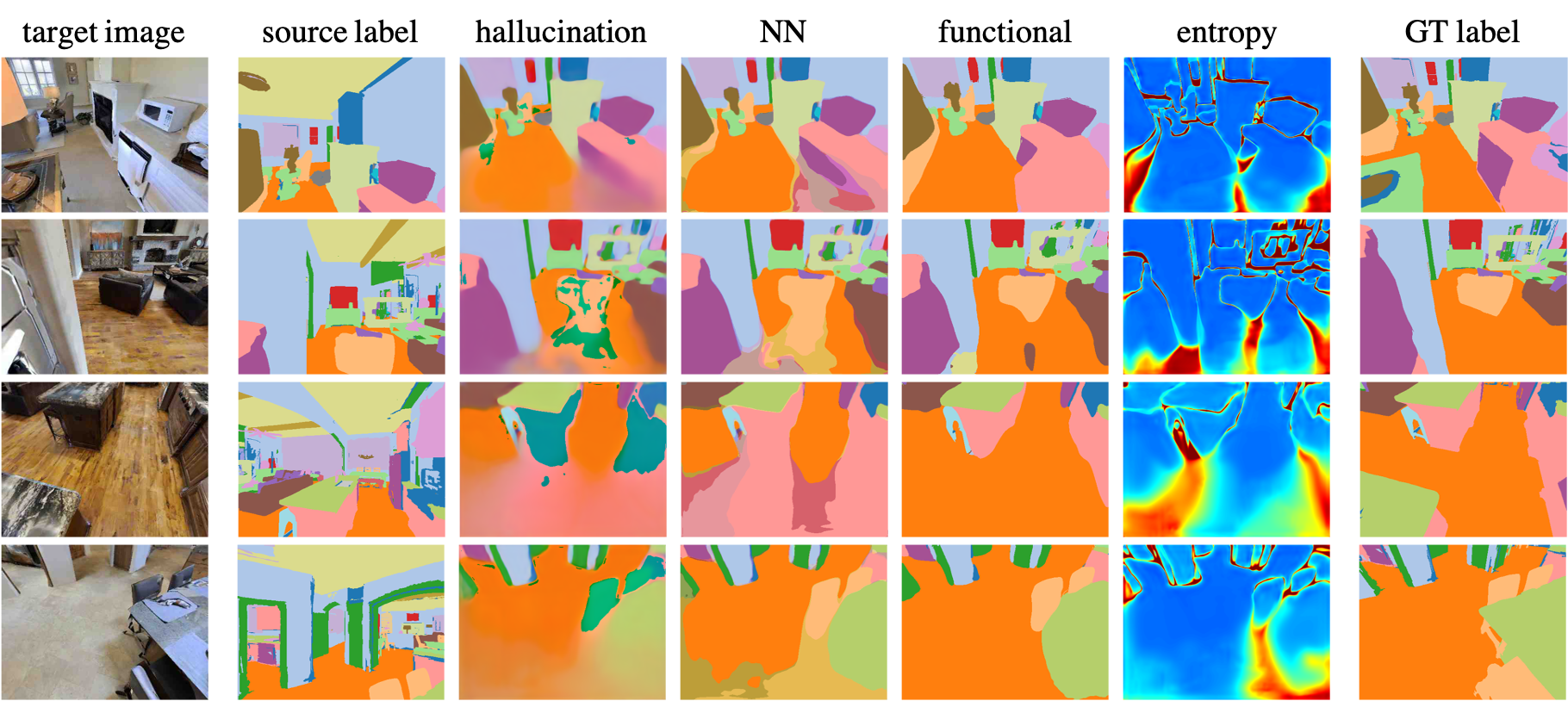}
\vspace{-0.2cm}
\caption{Effectiveness of the proposed functional transportation strategy. The target semantic images (3rd) are hallucinated from the source counterparts (2nd), which are decoded into semantic labels using nearest neighbor search (4th) or the proposed functional strategy (5th) with uncertainties (6th).}
\vspace{-0.3cm}
\label{fig:func-hallucinate}
\end{figure}

\vspace{-0.1cm}
\subsection{Functional label hallucination}
\vspace{-0.1cm}

Given the trained $\psi$, we can hallucinate the target semantic images for $x^\tau_i$'s, \textit{i.e.}, $\hat{y}^\tau_i=\psi(x_V; x_K,x_Q)$, by setting $x_Q=x^\tau_i$, $x_K=x^s_i$ and $x_V=y^s_i$.
We can then convert the hallucinated semantic images into semantic labels (integers) via nearest neighbor search in the color space to adapt a semantic segmentation network to the target domains.
However, the converted labels sometimes could be wrong due to noise in the predicted color (see Fig.~\ref{fig:func-hallucinate}).

To resolve the ambiguities, we propose the following functional label hallucination by treating $\psi(\cdot; x^s_i, x^\tau_i)$ as the functional representation of an unknown mapping $T(x^s_i, x^\tau_i): \Omega_s \rightarrow \Omega_\tau$ conditioned on the color images $x^s_i, x^\tau_i$. Here $\Omega_s,\Omega_\tau$ represent the source and target image domains/grids.
According to \cite{ovsjanikov2012functional}, if $T$ is a bijective mapping between $\Omega_s$ and $\Omega_\tau$, the actual mapping $T$ can then be recovered from $\psi(\cdot; x^s_i, x^\tau_i)$ by checking its output of indicator functions of the elements in $\Omega_s$.
However, recovering the underlying unknown mapping $T$ is unnecessary in our scenario, and, indeed, we do not have any constraints that $T$ is bijective.
Instead, we utilize the functional representation $\psi(\cdot; x^s_i, x^\tau_i)$ to find regions in $\Omega_\tau$ that share the same label with those in $\Omega_s$. 
Let $\mathbf{1}_{y^s_i=c}$ be the indicator function of the regions that are classified as class $c$, then $\hat{y}^{\tau}_{ic} = \psi(\mathbf{1}_{y^s_i=c}; x^s_i, x^\tau_i)$ indicates the regions of class $c$ in $\Omega_\tau$. And the hallucinated labels can be written as:
\begin{equation}
    \hat{y}^{\tau}_i = softmax(\psi(\mathbf{1}_{y^s_i=1}; x^s_i, x^\tau_i), ..., \psi(\mathbf{1}_{y^s_i=C}; x^s_i, x^\tau_i))
    \label{eq:func-decoding}
\end{equation}
with $C$ the number of semantic classes within the datasets, and now the hallucinated target view labels $\hat{y}^\tau_i$ represent the probabilistic distributions over the $C$ classes for each pixel.

{\bf Adapting to target domains.} 
With the functional transportation strategy,
we can avoid performing a nearest neighbor search in the color space, which improves the accuracy of the generated labels even when the hallucinated color is noisy (see Fig.~\ref{fig:func-hallucinate}).
Moreover, the soft probabilistic labels are more suitable for adapting the semantic segmentation network $\phi$ to the target domains, avoiding errors of hard labels when the hallucination is of low confidence.
We simply finetune $\phi$ for each target domain using the target dataset $\D^\tau = \{(x^\tau_i,\hat{y}^\tau_i)\}$ augmented with the soft labels:
\begin{equation}
    \loss^{\phi} = \sum_i\mathbb{H}(\hat{y}^\tau_i, \phi(x^\tau_i))
    \label{eq:task-loss}
\end{equation}
where $\mathbb{H}$ is the cross-entropy. 
Next, we introduce the evaluation benchmark and the experiments.

\begin{figure}[!t]
   \centering
   \includegraphics[width=0.92\linewidth]{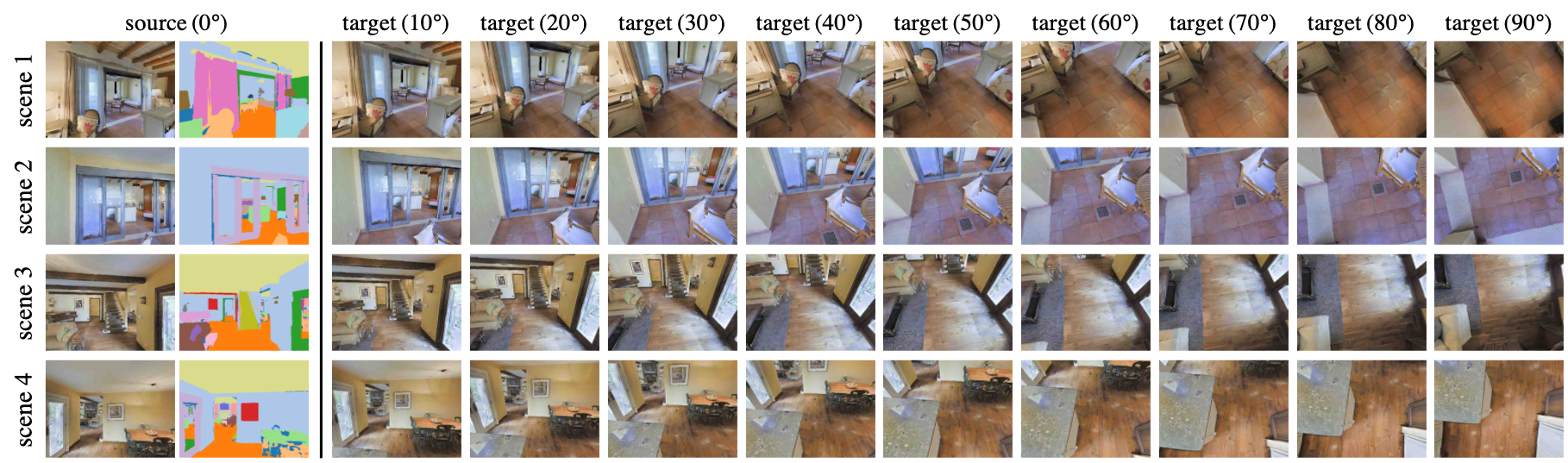}
\vspace{-0.2cm}
\caption{Samples from the proposed dataset (one source and nine target domains) for benchmarking unsupervised domain adaptation methods on viewpoint change in semantic segmentation.}
\vspace{-0.2cm}
\label{fig:datasets}
\end{figure}

\vspace{-0.2cm}
\section{Experiments}\label{sec:experiments}

\vspace{-0.1cm}
\subsection{Data generation}\label{sec:datagen}
\vspace{-0.1cm}

Due to the lack of benchmarks for evaluating UDA methods on viewpoint change, we propose a new dataset whose source and target domains are generated by varying camera elevation and viewing angles.
Moreover, we explicitly control the viewpoint changes, such that we can quantitatively assess the adaptation performance with respect to the degree of domain gaps.
We resort to simulation for data collection since 1) it is much easier to obtain semantic segmentation ground-truth in simulation; 2) the degree of the domain gap caused by viewpoint change is more controllable; and 3) it is more friendly to the personnel who is in charge of the data collection given the pandemic.

Furthermore, we maximize the realism of the generated data by employing the Matterport3D dataset \cite{chang2017matterport3d}, which contains 90 building-scale real-world scenes with pixel-wise semantic annotations\footnote{we completed the Terms of Use agreement form and obtained consent from the creators, and the data does not contain any personally identifiable information}. The scenes from Matterport3D are then imported into the Habitat simulation platform \cite{savva2019habitat} for our data generation.
Specifically, we first randomly sample two states in the scene, with one state (the position and yaw angle\footnote{orientation within a horizontal plane, similarly, the pitch angle is the orientation within a vertical plane} of a virtual camera) representing the starting state, and the other the end state.
We then perform collision-free path planning between these two states. The resulted path is accepted if it has a length larger than 15 path points, and images at each point along the path are collected.
To synthesize the domain gaps,
we set the pitch angle of the virtual camera to $0^\circ$ for collecting the source domain videos (annotations), which resembles the working viewpoint for semantic segmentation networks trained on existing scene parsing datasets \cite{silberman11indoor,song2015sun,zhou2017scene}.
Moreover, we increase the pitch angle of the virtual camera by $10^\circ$ (up to $90^\circ$) for collecting target domain videos (no annotations), which results in 9 different target domains.
For each domain, we collect 13,500 training images and 2,700 test images with resolution 384$\times$512.
Please see Fig.~\ref{fig:datasets} for samples from the collected datasets.

\vspace{-0.1cm}
\subsection{Implementation details.}\label{sec:implementation}
\vspace{-0.1cm}

We adapt the UNet structure \cite{ronneberger2015u} with reduced capacity and layernorm activation to construct the feed-forward networks $\ffn_Q$ and $\ffn_K$.
Similar to \cite{yan2020sttn}, $W$ is a convolutional layer with kernel size 1$\times$1, $\ffn_{V1},\ffn_{V2}$ consist of one and two convolutional layers respectively. Both $\ffn_{V1}$ and $\ffn_{V2}$ use leakyrelu as the activation function.
Our view transformation network contains $L=8$ attention modules. Training of the view transformation network $\psi$ is carried out on eight Nvidia V100 GPUs, with batch size 16.
We use the Adam optimizer with an initial learning rate of 1e-4 and momentums of 0.9 and 0.999. The training converges after 10 epochs.
We use the DeepLabv2 \cite{chen2017deeplab} with the ResNet101 backbone as the semantic segmentation network $\phi$, which is initialized with the pre-trained weights on ImageNet \cite{luo2019taking,zhang2019category,yang2020fda,li2020content,vu2019advent}. Soft labels for each target view $\tau$ are hallucinated using Eq.~\eqref{eq:func-decoding}. Moreover, the semantic segmentation networks $\phi^\tau$ for each target domain are trained using Eq.~\eqref{eq:task-loss} with the Adam optimizer, with a batch size of 6 and an initial learning rate of 7.5e-5. The learning rate is then halved after 10 and 15 epochs. The training converges at 25 epochs. To have a fair comparison with the state-of-the-art domain adaptation methods that adapt from a single source domain to a single target domain, we also train the segmentation network for each target domain separately. We use mean intersection-over-union (mIoU) as the metric.

\begin{table}
  \caption{Ablation study on different inductive biases for zero-shot semantic image hallucination. Numbers are the mIoUs of the hallucinated semantic labels on the training set of each target domain.}
  \vspace{-0.1cm}
  \label{tab:ablation-inductive}
  \centering
  \small
  \setlength{\tabcolsep}{4pt}
  \begin{tabular}{lccccccccc}
    \toprule
    & \multicolumn{9}{c}{Target Domains} \\
    \cmidrule(lr){2-10}
    Method & $10^{\circ}$ & $20^{\circ}$ & $30^{\circ}$ & $40^{\circ}$ & $50^{\circ}$ & $60^{\circ}$ & $70^{\circ}$ & $80^{\circ}$ & $90^{\circ}$ \\
    \midrule
    UNet     & 49.76 & 28.19 & 13.69 & 9.26 & 6.56 & 4.71 & 2.59 & 1.63 & 1.28 \\
    Flow     & 33.04 & 27.59 & 22.72 & 19.36 & 17.02 & 14.21 & 11.55 & 9.67 & 8.34 \\
    RAFT \cite{teed2020raft} & 70.62 & 61.25 & 53.92 & 42.54 & 18.17 & 9.36 & 7.57 & 6.24 & 5.58 \\
    3D     & 28.16 & 22.12 & 18.35 & 15.80 & 13.14 & 11.22 & 9.20 & 6.61 & 2.86 \\
    ADeLA (single) & 54.85 & 46.29 & 42.66 & 37.75 & 27.71 & 21.33 & 14.18 & 8.69 & 4.17 \\
    ADeLA (multiple) & 48.42 & 41.87 & 35.73 & 30.39 & 24.11 & 17.40 & 11.79 & 8.82 & 7.34 \\
    \midrule
    UNet+F     & 73.62 & 49.07 & 27.12 & 20.08 & 16.48 & 13.68 & 11.61 & 9.79 & 8.53 \\
    ADeLA (single)+F & 70.07 & \bf 67.63 & \bf 58.62 & \bf 54.33 & \bf 47.45 & \bf 37.81 & \bf 28.39 & \bf 19.78 & \bf 15.17 \\
    ADeLA (multiple)+F & \bf 75.75 & 66.29 & 57.45 & 49.57 & 40.38 & 30.00 & 20.96 & 15.44 & 12.60 \\
    \bottomrule
  \end{tabular}
  \vspace{-0.2cm}
\end{table}

\vspace{-0.1cm}
\subsection{Ablation study}
\vspace{-0.1cm}

{\bf Effectiveness of the proposed inductive bias.} 
Qualitative comparisons in Fig.~\ref{fig:attention-is-effective} show that the proposed inductive bias and the architecture facilitate zero-shot semantic image hallucination.
In Tab.~\ref{tab:ablation-inductive} we quantitatively confirm its effectiveness and check how it extends across different levels of view-dependent domain gaps.
Besides the color transformation bias (UNet), we also test the inductive biases introduced by explicitly modeling the dense 2D correspondence ("Flow") and by explicitly modeling the image formation process in 3D ("3D").
For "Flow," we adapt the architecture of RAFT \cite{teed2020raft} and train it to estimate the flow that reconstructs the target color image from the source, and use the flow for warping the semantic labels.
For "3D", we adapt the state-of-the-art single view synthesis framework \cite{wiles2020synsin}, and supply it with ground-truth camera poses for semantic image synthesis.
We report the performance of our method under two settings. One is the single source to single target setting (ADeLA (single)), the other is the single source to multiple targets setting (ADeLA (multiple)).
The labels for ADeLA (single) and ADeLA (multiple) are generated using nearest neighbor search.
We also report the score of the warped labels using the fully supervised RAFT model for reference.

We can make the following observations: 1) UNet (color transformation) does not work at large viewpoint change. 2) the 2D dense correspondence inductive bias ("Flow") works better for large viewpoint change, which verifies our proposal for biasing towards transportation. However, the comparison between "Flow" and "RAFT" shows that the spatial correspondence learned from color images can be erroneous, so "Flow" is much worse than "RAFT" at small viewpoint changes. Moreover, "RAFT" is worse than "Flow" at large viewpoint changes, which indicates that the exact dense correspondence may not be suitable for semantic label hallucination.
3) The 3D inductive bias ("3D") does not perform well since the learned 3D representation from color images does not generalize to semantic images.
4) Our model performs well across all target domains, due to the proposed spatial transportation bias, and the capability to hallucinate beyond exact correspondence.

Moreover, we show the quality of the semantic labels hallucinated using the proposed functional label hallucination strategy ("UNet+F," "ADeLA(single)+F," "ADeLA(multiple)+F"). 
As seen in Tab.~\ref{tab:ablation-inductive} (bottom), functional hallucination significantly improves the performance of UNet and our models, demonstrating its effectiveness in resolving the ambiguities in the hallucinated semantic images.
Note, "Flow" and RAFT warp labels with explicit dense correspondence, thus they are unable to take advantage of the functional strategy. The same observation holds for "3D", whose 3D representation learned with color images does not generalize even if supplied with ground-truth camera poses.

\begin{figure}[!t]
   \centering
   \includegraphics[width=0.81\linewidth]{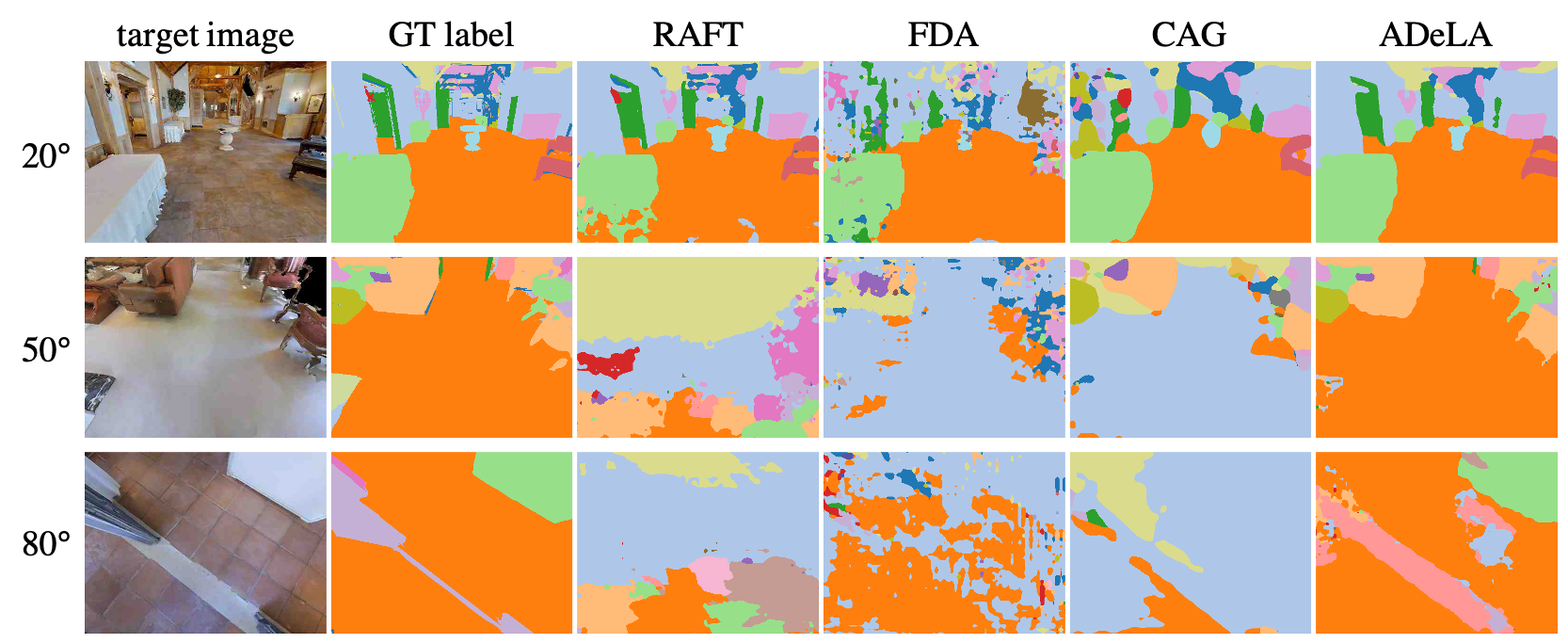}
\vspace{-0.3cm}
\caption{Qualitative comparison to competing methods on different target domains.}
\vspace{-0.2cm}
\label{fig:visual-results}
\end{figure}

\begin{table}[!t]
  \caption{Quantitative comparison to state-of-the-art methods on the proposed benchmark. Numbers are mIoU scores of the adapted semantic segmentation networks on the test set of each target domain.}
  \vspace{-0.1cm}
  \label{tab:benchmarking}
  \centering
  \small
  \begin{tabular}{lccccccccc}
    \toprule
    & \multicolumn{9}{c}{Target Domains} \\
    \cmidrule(lr){2-10}
    Method & $10^{\circ}$ & $20^{\circ}$ & $30^{\circ}$ & $40^{\circ}$ & $50^{\circ}$ & $60^{\circ}$ & $70^{\circ}$ & $80^{\circ}$ & $90^{\circ}$ \\
    \midrule
    Baseline & 27.90 & 20.24 & 12.22 & 7.63 & 4.41 & 2.53 & 1.80 & 1.53 & 1.53  \\
    RAFT \cite{teed2020raft} & 29.46 & \underline{26.74} & \underline{24.15} & 16.92 & \underline{8.66} & 4.54 & 3.35 & 2.82 & 2.41 \\
    MFNet \cite{zhao2020maskflownet} & 29.20 & 24.53 & 13.16 & 6.65 & 4.76 & 3.95 & 3.00 & 2.81 & 2.48 \\
    DICL \cite{wang2020displacement} & 29.62 & 26.45 & 22.01 & \underline{17.75} & 6.25 & 4.44 & 3.40 & 2.75 & 2.49 \\
    ProDA \cite{zhang2021prototypical} & 25.26 & 19.35 & 12.03 & 7.46 & 4.39 & 1.74 & 1.12 & 0.83 & 0.77 \\
    CLAN \cite{luo2019taking} & 28.22 & 21.21 & 13.17 & 7.53 & 4.37 & 2.66 & 2.02 & 1.74 & 1.73 \\
    CAG \cite{zhang2019category} & 27.17 & 22.22 & 15.05 & 8.57 & 5.24 & 2.43 & 1.83 & 1.50 & 1.54 \\
    FDA \cite{yang2020fda} & \bf 37.80 & 23.34 & 12.33 & 6.69 & 3.58 & 2.15 & 1.56 & 1.67 & 1.32 \\
    LTIR \cite{kim2020learning} & 26.22 & 20.50 & 13.43 & 6.16 & 3.90 & 2.09 & 1.82 & 1.65 & 1.66 \\
    CCM \cite{li2020content} & 28.26 & 19.48 & 10.56 & 4.92 & 2.78 & 1.50 & 1.14 & 0.95 & 0.90 \\
    Advent \cite{vu2019advent} & 11.38 & 7.93 & 4.98 & 3.28 & 2.54 & 2.16 & 1.60 & 1.52 & 1.49 \\
    Intrada \cite{pan2020unsupervised} & 10.16 & 7.84 & 6.13 & 4.08 & 2.67 & 1.98 & 1.58 & 1.67 & 0.93 \\
    AppFlow \cite{zhou2016view} & 14.73 & 12.91 & 10.46 & 8.43 & 7.30 & \underline{5.68} & 4.66 & \underline{3.87} & \underline{3.39} \\
    Synsin \cite{wiles2020synsin} & 14.43 & 11.44 & 9.15 & 8.29 & 6.24 & 5.43 & \underline{4.67} & 3.36 & 1.88\\
    UNet     & 30.51 & 22.36 & 12.26 & 8.44 & 6.58 & 4.80 & 3.81 & 2.98 & 2.22 \\
    ADeLA    & \underline{31.19} & \bf 27.66 & \bf 24.31 & \bf 20.71 & \bf 15.88 & \bf 12.28 & \bf 8.27 & \bf 7.16 & \bf 5.03 \\
    \bottomrule
  \end{tabular}
  \vspace{-0.2cm}
\end{table}

\vspace{-0.1cm}
\subsection{Benchmarking}
\vspace{-0.1cm}

We carry out an extensive study of state-of-the-art methods on unsupervised domain adaptation for viewpoint changes in semantic segmentation \cite{zhang2021prototypical,luo2019taking,zhang2019category,yang2020fda,kim2020learning,li2020content,vu2019advent,pan2020unsupervised}. 
Among those methods, \cite{zhang2021prototypical,li2020content} focus on self-training, \cite{luo2019taking,zhang2019category,pan2020unsupervised} perform class-wise and curriculum domain alignment, and \cite{yang2020fda,vu2019advent,kim2020learning} align domains in the image/output space.
We also experiment with three best performing dense correspondence estimation methods \cite{teed2020raft,zhao2020maskflownet,wang2020displacement}, and two single view synthesis methods \cite{zhou2016view,wiles2020synsin} to generate target view labels to help adapt the semantic segmentation networks.
All methods are re-trained on the training sets of the proposed benchmark, and tested on the test sets of the nine target domains.
Our method consistently achieves positive adaptation gain and performs much better than the other methods at large viewpoint changes.
Note FDA \cite{yang2020fda} performs better on the target domain of $10^\circ$ ( small gap) due to its strong style randomization mechanism. However, our method outperforms FDA by a significant margin on the remaining target domains even without any data augmentation in adapting the semantic segmentation network. See Fig.~\ref{fig:visual-results} for visual comparisons.

\vspace{-0.1cm}
\section{Discussion}\label{sec:discussion}
\vspace{-0.1cm}

We tackle UDA for viewpoint change in the semantic segmentation task.
Extensive experiments on the new benchmark collected with controlled domain gaps caused by changing views demonstrate the effectiveness of the proposed inductive bias for zero-shot view hallucination, strengthened by the proposed functional transportation strategy, in reducing the domain gaps. 
Our experiments also verify that aligning statistics of domains in a shared space could be counter-productive due to the content shift across viewing angles.
Our method achieves higher adaptation gains, especially in the regime of large viewpoint changes.
However, the adaptation gain of our method also decreases toward the extreme case where the viewing angles are perpendicular.
Like many visual perception methods, there is a potential ethical and societal concern when applied to downstream applications. In our code release, we will explicitly specify allowable uses of our system with appropriate licenses.
In the future, we would like to explore how the temporal information can be utilized to reduce further the domain gaps caused by extreme views.

{\small
\bibliographystyle{ieee_fullname}
\bibliography{egbib}

\begin{thebibliography}{10}\itemsep=-1pt

\bibitem{bak2018domain}
Slawomir Bak, Peter Carr, and Jean-Francois Lalonde.
\newblock Domain adaptation through synthesis for unsupervised person
  re-identification.
\newblock In {\em Proceedings of the European Conference on Computer Vision
  (ECCV)}, pages 189--205, 2018.

\bibitem{baktashmotlagh2013unsupervised}
Mahsa Baktashmotlagh, Mehrtash~T Harandi, Brian~C Lovell, and Mathieu Salzmann.
\newblock Unsupervised domain adaptation by domain invariant projection.
\newblock In {\em Proceedings of the IEEE International Conference on Computer
  Vision}, pages 769--776, 2013.

\bibitem{chang2017matterport3d}
Angel Chang, Angela Dai, Thomas Funkhouser, Maciej Halber, Matthias Niessner,
  Manolis Savva, Shuran Song, Andy Zeng, and Yinda Zhang.
\newblock Matterport3d: Learning from rgb-d data in indoor environments.
\newblock {\em arXiv preprint arXiv:1709.06158}, 2017.

\bibitem{chang2019all}
Wei-Lun Chang, Hui-Po Wang, Wen-Hsiao Peng, and Wei-Chen Chiu.
\newblock All about structure: Adapting structural information across domains
  for boosting semantic segmentation.
\newblock In {\em Proceedings of the IEEE Conference on Computer Vision and
  Pattern Recognition}, pages 1900--1909, 2019.

\bibitem{chen2017deeplab}
Liang-Chieh Chen, George Papandreou, Iasonas Kokkinos, Kevin Murphy, and Alan~L
  Yuille.
\newblock Deeplab: Semantic image segmentation with deep convolutional nets,
  atrous convolution, and fully connected crfs.
\newblock {\em IEEE transactions on pattern analysis and machine intelligence},
  40(4):834--848, 2017.

\bibitem{choi2019extreme}
Inchang Choi, Orazio Gallo, Alejandro Troccoli, Min~H Kim, and Jan Kautz.
\newblock Extreme view synthesis.
\newblock In {\em Proceedings of the IEEE/CVF International Conference on
  Computer Vision}, pages 7781--7790, 2019.

\bibitem{combes2020domain}
Remi Tachet~des Combes, Han Zhao, Yu-Xiang Wang, and Geoff Gordon.
\newblock Domain adaptation with conditional distribution matching and
  generalized label shift.
\newblock {\em arXiv preprint arXiv:2003.04475}, 2020.

\bibitem{coors2019nova}
Benjamin Coors, Alexandru~Paul Condurache, and Andreas Geiger.
\newblock Nova: Learning to see in novel viewpoints and domains.
\newblock In {\em 2019 International Conference on 3D Vision (3DV)}, pages
  116--125. IEEE, 2019.

\bibitem{csurka2017domain}
Gabriela Csurka.
\newblock A comprehensive survey on domain adaptation for visual applications.
\newblock In {\em Domain adaptation in computer vision applications}, pages
  1--35. Springer, 2017.

\bibitem{deng2018image}
Weijian Deng, Liang Zheng, Qixiang Ye, Guoliang Kang, Yi Yang, and Jianbin
  Jiao.
\newblock Image-image domain adaptation with preserved self-similarity and
  domain-dissimilarity for person re-identification.
\newblock In {\em Proceedings of the IEEE conference on computer vision and
  pattern recognition}, pages 994--1003, 2018.

\bibitem{di2020sceneadapt}
Daniele Di~Mauro, Antonino Furnari, Giuseppe Patan{\`e}, Sebastiano Battiato,
  and Giovanni~Maria Farinella.
\newblock Sceneadapt: Scene-based domain adaptation for semantic segmentation
  using adversarial learning.
\newblock {\em Pattern Recognition Letters}, 2020.

\bibitem{dong2020cscl}
Jiahua Dong, Yang Cong, Gan Sun, Yuyang Liu, and Xiaowei Xu.
\newblock Cscl: Critical semantic-consistent learning for unsupervised domain
  adaptation.
\newblock In {\em European Conference on Computer Vision}, pages 745--762.
  Springer, 2020.

\bibitem{du2019ssf}
Liang Du, Jingang Tan, Hongye Yang, Jianfeng Feng, Xiangyang Xue, Qibao Zheng,
  Xiaoqing Ye, and Xiaolin Zhang.
\newblock Ssf-dan: Separated semantic feature based domain adaptation network
  for semantic segmentation.
\newblock In {\em Proceedings of the IEEE/CVF International Conference on
  Computer Vision}, pages 982--991, 2019.

\bibitem{fernando2013unsupervised}
Basura Fernando, Amaury Habrard, Marc Sebban, and Tinne Tuytelaars.
\newblock Unsupervised visual domain adaptation using subspace alignment.
\newblock In {\em Proceedings of the IEEE international conference on computer
  vision}, pages 2960--2967, 2013.

\bibitem{flynn2019deepview}
John Flynn, Michael Broxton, Paul Debevec, Matthew DuVall, Graham Fyffe, Ryan
  Overbeck, Noah Snavely, and Richard Tucker.
\newblock Deepview: View synthesis with learned gradient descent.
\newblock In {\em Proceedings of the IEEE/CVF Conference on Computer Vision and
  Pattern Recognition}, pages 2367--2376, 2019.

\bibitem{fu2019self}
Yang Fu, Yunchao Wei, Guanshuo Wang, Yuqian Zhou, Honghui Shi, and Thomas~S
  Huang.
\newblock Self-similarity grouping: A simple unsupervised cross domain
  adaptation approach for person re-identification.
\newblock In {\em Proceedings of the IEEE/CVF International Conference on
  Computer Vision}, pages 6112--6121, 2019.

\bibitem{geng2011daml}
Bo Geng, Dacheng Tao, and Chao Xu.
\newblock Daml: Domain adaptation metric learning.
\newblock {\em IEEE Transactions on Image Processing}, 20(10):2980--2989, 2011.

\bibitem{glorot2011domain}
Xavier Glorot, Antoine Bordes, and Yoshua Bengio.
\newblock Domain adaptation for large-scale sentiment classification: A deep
  learning approach.
\newblock In {\em Proceedings of the 28th international conference on machine
  learning (ICML-11)}, pages 513--520, 2011.

\bibitem{gong2019dlow}
Rui Gong, Wen Li, Yuhua Chen, and Luc~Van Gool.
\newblock Dlow: Domain flow for adaptation and generalization.
\newblock In {\em Proceedings of the IEEE Conference on Computer Vision and
  Pattern Recognition}, pages 2477--2486, 2019.

\bibitem{hoffman2018cycada}
Judy Hoffman, Eric Tzeng, Taesung Park, Jun-Yan Zhu, Phillip Isola, Kate
  Saenko, Alexei Efros, and Trevor Darrell.
\newblock Cycada: Cycle-consistent adversarial domain adaptation.
\newblock In {\em Proceedings of the 35th International Conference on Machine
  Learning}, 2018.

\bibitem{huang2020contextual}
Jiaxing Huang, Shijian Lu, Dayan Guan, and Xiaobing Zhang.
\newblock Contextual-relation consistent domain adaptation for semantic
  segmentation.
\newblock In {\em European Conference on Computer Vision}, pages 705--722.
  Springer, 2020.

\bibitem{kim2020learning}
Myeongjin Kim and Hyeran Byun.
\newblock Learning texture invariant representation for domain adaptation of
  semantic segmentation.
\newblock In {\em Proceedings of the IEEE/CVF Conference on Computer Vision and
  Pattern Recognition}, pages 12975--12984, 2020.

\bibitem{kumar2018co}
Abhishek Kumar, Prasanna Sattigeri, Kahini Wadhawan, Leonid Karlinsky, Rogerio
  Feris, Bill Freeman, and Gregory Wornell.
\newblock Co-regularized alignment for unsupervised domain adaptation.
\newblock In {\em Advances in Neural Information Processing Systems}, pages
  9345--9356, 2018.

\bibitem{li2020content}
Guangrui Li, Guoliang Kang, Wu Liu, Yunchao Wei, and Yi Yang.
\newblock Content-consistent matching for domain adaptive semantic
  segmentation.
\newblock In {\em European Conference on Computer Vision}, pages 440--456.
  Springer, 2020.

\bibitem{li2019bidirectional}
Yunsheng Li, Lu Yuan, and Nuno Vasconcelos.
\newblock Bidirectional learning for domain adaptation of semantic
  segmentation.
\newblock In {\em Proceedings of the IEEE Conference on Computer Vision and
  Pattern Recognition}, pages 6936--6945, 2019.

\bibitem{liu2017unsupervised}
Ming-Yu Liu, Thomas Breuel, and Jan Kautz.
\newblock Unsupervised image-to-image translation networks.
\newblock In {\em Advances in neural information processing systems}, pages
  700--708, 2017.

\bibitem{long2015learning}
Mingsheng Long, Yue Cao, Jianmin Wang, and Michael Jordan.
\newblock Learning transferable features with deep adaptation networks.
\newblock In {\em International Conference on Machine Learning}, pages 97--105,
  2015.

\bibitem{luo2019significance}
Yawei Luo, Ping Liu, Tao Guan, Junqing Yu, and Yi Yang.
\newblock Significance-aware information bottleneck for domain adaptive
  semantic segmentation.
\newblock In {\em Proceedings of the IEEE International Conference on Computer
  Vision}, pages 6778--6787, 2019.

\bibitem{luo2019taking}
Yawei Luo, Liang Zheng, Tao Guan, Junqing Yu, and Yi Yang.
\newblock Taking a closer look at domain shift: Category-level adversaries for
  semantics consistent domain adaptation.
\newblock In {\em Proceedings of the IEEE Conference on Computer Vision and
  Pattern Recognition}, pages 2507--2516, 2019.

\bibitem{motiian2017unified}
Saeid Motiian, Marco Piccirilli, Donald~A Adjeroh, and Gianfranco Doretto.
\newblock Unified deep supervised domain adaptation and generalization.
\newblock In {\em Proceedings of the IEEE International Conference on Computer
  Vision}, pages 5715--5725, 2017.

\bibitem{ovsjanikov2012functional}
Maks Ovsjanikov, Mirela Ben-Chen, Justin Solomon, Adrian Butscher, and Leonidas
  Guibas.
\newblock Functional maps: a flexible representation of maps between shapes.
\newblock {\em ACM Transactions on Graphics (TOG)}, 31(4):1--11, 2012.

\bibitem{pan2020unsupervised}
Fei Pan, Inkyu Shin, Francois Rameau, Seokju Lee, and In~So Kweon.
\newblock Unsupervised intra-domain adaptation for semantic segmentation
  through self-supervision.
\newblock In {\em Proceedings of the IEEE/CVF Conference on Computer Vision and
  Pattern Recognition}, pages 3764--3773, 2020.

\bibitem{patel2015visual}
Vishal~M Patel, Raghuraman Gopalan, Ruonan Li, and Rama Chellappa.
\newblock Visual domain adaptation: A survey of recent advances.
\newblock {\em IEEE signal processing magazine}, 32(3):53--69, 2015.

\bibitem{peng2017visda}
Xingchao Peng, Ben Usman, Neela Kaushik, Judy Hoffman, Dequan Wang, and Kate
  Saenko.
\newblock Visda: The visual domain adaptation challenge.
\newblock {\em arXiv preprint arXiv:1710.06924}, 2017.

\bibitem{peng2018syn2real}
Xingchao Peng, Ben Usman, Kuniaki Saito, Neela Kaushik, Judy Hoffman, and Kate
  Saenko.
\newblock Syn2real: A new benchmark forsynthetic-to-real visual domain
  adaptation.
\newblock {\em arXiv preprint arXiv: 1806.09755}, 2018.

\bibitem{rahmani2017learning}
Hossein Rahmani, Ajmal Mian, and Mubarak Shah.
\newblock Learning a deep model for human action recognition from novel
  viewpoints.
\newblock {\em IEEE transactions on pattern analysis and machine intelligence},
  40(3):667--681, 2017.

\bibitem{ronneberger2015u}
Olaf Ronneberger, Philipp Fischer, and Thomas Brox.
\newblock U-net: Convolutional networks for biomedical image segmentation.
\newblock In {\em International Conference on Medical image computing and
  computer-assisted intervention}, pages 234--241. Springer, 2015.

\bibitem{saenko2010adapting}
Kate Saenko, Brian Kulis, Mario Fritz, and Trevor Darrell.
\newblock Adapting visual category models to new domains.
\newblock In {\em European conference on computer vision}, pages 213--226.
  Springer, 2010.

\bibitem{saito2017asymmetric}
Kuniaki Saito, Yoshitaka Ushiku, and Tatsuya Harada.
\newblock Asymmetric tri-training for unsupervised domain adaptation.
\newblock In {\em Proceedings of the 34th International Conference on Machine
  Learning-Volume 70}, pages 2988--2997. JMLR. org, 2017.

\bibitem{sankaranarayanan2018learning}
Swami Sankaranarayanan, Yogesh Balaji, Arpit Jain, Ser Nam~Lim, and Rama
  Chellappa.
\newblock Learning from synthetic data: Addressing domain shift for semantic
  segmentation.
\newblock In {\em Proceedings of the IEEE Conference on Computer Vision and
  Pattern Recognition}, pages 3752--3761, 2018.

\bibitem{savva2019habitat}
Manolis Savva, Abhishek Kadian, Oleksandr Maksymets, Yili Zhao, Erik Wijmans,
  Bhavana Jain, Julian Straub, Jia Liu, Vladlen Koltun, Jitendra Malik, et~al.
\newblock Habitat: A platform for embodied ai research.
\newblock In {\em Proceedings of the IEEE/CVF International Conference on
  Computer Vision}, pages 9339--9347, 2019.

\bibitem{sener2016learning}
Ozan Sener, Hyun~Oh Song, Ashutosh Saxena, and Silvio Savarese.
\newblock Learning transferrable representations for unsupervised domain
  adaptation.
\newblock In {\em Advances in Neural Information Processing Systems}, pages
  2110--2118, 2016.

\bibitem{shu2018dirt}
Rui Shu, Hung~H Bui, Hirokazu Narui, and Stefano Ermon.
\newblock A dirt-t approach to unsupervised domain adaptation.
\newblock In {\em Proc. 6th International Conference on Learning
  Representations}, 2018.

\bibitem{silberman11indoor}
N. Silberman and R. Fergus.
\newblock Indoor scene segmentation using a structured light sensor.
\newblock In {\em Proceedings of the International Conference on Computer
  Vision - Workshop on 3D Representation and Recognition}, 2011.

\bibitem{sitzmann2019scene}
Vincent Sitzmann, Michael Zollh{\"o}fer, and Gordon Wetzstein.
\newblock Scene representation networks: Continuous 3d-structure-aware neural
  scene representations.
\newblock {\em arXiv preprint arXiv:1906.01618}, 2019.

\bibitem{song2015sun}
Shuran Song, Samuel~P Lichtenberg, and Jianxiong Xiao.
\newblock Sun rgb-d: A rgb-d scene understanding benchmark suite.
\newblock In {\em Proceedings of the IEEE conference on computer vision and
  pattern recognition}, pages 567--576, 2015.

\bibitem{teed2020raft}
Zachary Teed and Jia Deng.
\newblock Raft: Recurrent all-pairs field transforms for optical flow.
\newblock In {\em European Conference on Computer Vision}, pages 402--419.
  Springer, 2020.

\bibitem{tsai2019domain}
Yi-Hsuan Tsai, Kihyuk Sohn, Samuel Schulter, and Manmohan Chandraker.
\newblock Domain adaptation for structured output via discriminative patch
  representations.
\newblock In {\em Proceedings of the IEEE/CVF International Conference on
  Computer Vision}, pages 1456--1465, 2019.

\bibitem{tucker2020single}
Richard Tucker and Noah Snavely.
\newblock Single-view view synthesis with multiplane images.
\newblock In {\em Proceedings of the IEEE/CVF Conference on Computer Vision and
  Pattern Recognition}, pages 551--560, 2020.

\bibitem{tzeng2017adversarial}
Eric Tzeng, Judy Hoffman, Kate Saenko, and Trevor Darrell.
\newblock Adversarial discriminative domain adaptation.
\newblock In {\em Proceedings of the IEEE conference on computer vision and
  pattern recognition}, pages 7167--7176, 2017.

\bibitem{vaswani2017attention}
Ashish Vaswani, Noam Shazeer, Niki Parmar, Jakob Uszkoreit, Llion Jones,
  Aidan~N Gomez, Lukasz Kaiser, and Illia Polosukhin.
\newblock Attention is all you need.
\newblock {\em arXiv preprint arXiv:1706.03762}, 2017.

\bibitem{vu2019advent}
Tuan-Hung Vu, Himalaya Jain, Maxime Bucher, Matthieu Cord, and Patrick
  P{\'e}rez.
\newblock Advent: Adversarial entropy minimization for domain adaptation in
  semantic segmentation.
\newblock In {\em Proceedings of the IEEE Conference on Computer Vision and
  Pattern Recognition}, pages 2517--2526, 2019.

\bibitem{wang2020displacement}
Jianyuan Wang, Yiran Zhong, Yuchao Dai, Kaihao Zhang, Pan Ji, and Hongdong Li.
\newblock Displacement-invariant matching cost learning for accurate optical
  flow estimation.
\newblock {\em arXiv preprint arXiv:2010.14851}, 2020.

\bibitem{wang2018deep}
Mei Wang and Weihong Deng.
\newblock Deep visual domain adaptation: A survey.
\newblock {\em Neurocomputing}, 312:135--153, 2018.

\bibitem{wang2020differential}
Zhonghao Wang, Mo Yu, Yunchao Wei, Rogerio Feris, Jinjun Xiong, Wen-mei Hwu,
  Thomas~S Huang, and Honghui Shi.
\newblock Differential treatment for stuff and things: A simple unsupervised
  domain adaptation method for semantic segmentation.
\newblock In {\em Proceedings of the IEEE/CVF Conference on Computer Vision and
  Pattern Recognition}, pages 12635--12644, 2020.

\bibitem{wiles2020synsin}
Olivia Wiles, Georgia Gkioxari, Richard Szeliski, and Justin Johnson.
\newblock Synsin: End-to-end view synthesis from a single image.
\newblock In {\em Proceedings of the IEEE/CVF Conference on Computer Vision and
  Pattern Recognition}, pages 7467--7477, 2020.

\bibitem{wu2018dcan}
Zuxuan Wu, Xintong Han, Yen-Liang Lin, Mustafa Gokhan~Uzunbas, Tom Goldstein,
  Ser Nam~Lim, and Larry~S Davis.
\newblock Dcan: Dual channel-wise alignment networks for unsupervised scene
  adaptation.
\newblock In {\em Proceedings of the European Conference on Computer Vision
  (ECCV)}, pages 518--534, 2018.

\bibitem{yang2020phase}
Yanchao Yang, Dong Lao, Ganesh Sundaramoorthi, and Stefano Soatto.
\newblock Phase consistent ecological domain adaptation.
\newblock In {\em Proceedings of the IEEE/CVF Conference on Computer Vision and
  Pattern Recognition}, pages 9011--9020, 2020.

\bibitem{yang2020fda}
Yanchao Yang and Stefano Soatto.
\newblock Fda: Fourier domain adaptation for semantic segmentation.
\newblock In {\em Proceedings of the IEEE/CVF Conference on Computer Vision and
  Pattern Recognition}, pages 4085--4095, 2020.

\bibitem{yi2017dualgan}
Zili Yi, Hao Zhang, Ping Tan, and Minglun Gong.
\newblock Dualgan: Unsupervised dual learning for image-to-image translation.
\newblock In {\em Proceedings of the IEEE international conference on computer
  vision}, pages 2849--2857, 2017.

\bibitem{zellinger2017central}
Werner Zellinger, Thomas Grubinger, Edwin Lughofer, Thomas Natschl{\"a}ger, and
  Susanne Saminger-Platz.
\newblock Central moment discrepancy (cmd) for domain-invariant representation
  learning.
\newblock {\em arXiv preprint arXiv:1702.08811}, 2017.

\bibitem{yan2020sttn}
Yanhong Zeng, Jianlong Fu, and Hongyang Chao.
\newblock Learning joint spatial-temporal transformations for video inpainting.
\newblock In {\em The Proceedings of the European Conference on Computer Vision
  (ECCV)}, 2020.

\bibitem{zhang2017view}
Pengfei Zhang, Cuiling Lan, Junliang Xing, Wenjun Zeng, Jianru Xue, and Nanning
  Zheng.
\newblock View adaptive recurrent neural networks for high performance human
  action recognition from skeleton data.
\newblock In {\em Proceedings of the IEEE International Conference on Computer
  Vision}, pages 2117--2126, 2017.

\bibitem{zhang2021prototypical}
Pan Zhang, Bo Zhang, Ting Zhang, Dong Chen, Yong Wang, and Fang Wen.
\newblock Prototypical pseudo label denoising and target structure learning for
  domain adaptive semantic segmentation.
\newblock {\em arXiv preprint arXiv:2101.10979}, 2, 2021.

\bibitem{zhang2019category}
Qiming Zhang, Jing Zhang, Wei Liu, and Dacheng Tao.
\newblock Category anchor-guided unsupervised domain adaptation for semantic
  segmentation.
\newblock {\em arXiv preprint arXiv:1910.13049}, 2019.

\bibitem{zhang2017curriculum}
Yang Zhang, Philip David, and Boqing Gong.
\newblock Curriculum domain adaptation for semantic segmentation of urban
  scenes.
\newblock In {\em Proceedings of the IEEE International Conference on Computer
  Vision}, pages 2020--2030, 2017.

\bibitem{zhao2020maskflownet}
Shengyu Zhao, Yilun Sheng, Yue Dong, Eric~I Chang, Yan Xu, et~al.
\newblock Maskflownet: Asymmetric feature matching with learnable occlusion
  mask.
\newblock In {\em Proceedings of the IEEE/CVF Conference on Computer Vision and
  Pattern Recognition}, pages 6278--6287, 2020.

\bibitem{zhou2017scene}
Bolei Zhou, Hang Zhao, Xavier Puig, Sanja Fidler, Adela Barriuso, and Antonio
  Torralba.
\newblock Scene parsing through ade20k dataset.
\newblock In {\em Proceedings of the IEEE Conference on Computer Vision and
  Pattern Recognition}, 2017.

\bibitem{zhou2018stereo}
Tinghui Zhou, Richard Tucker, John Flynn, Graham Fyffe, and Noah Snavely.
\newblock Stereo magnification: Learning view synthesis using multiplane
  images.
\newblock {\em arXiv preprint arXiv:1805.09817}, 2018.

\bibitem{zhou2016view}
Tinghui Zhou, Shubham Tulsiani, Weilun Sun, Jitendra Malik, and Alexei~A Efros.
\newblock View synthesis by appearance flow.
\newblock In {\em European conference on computer vision}, pages 286--301.
  Springer, 2016.

\bibitem{zhu2017unpaired}
Jun-Yan Zhu, Taesung Park, Phillip Isola, and Alexei~A Efros.
\newblock Unpaired image-to-image translation using cycle-consistent
  adversarial networks.
\newblock In {\em Proceedings of the IEEE international conference on computer
  vision}, pages 2223--2232, 2017.

\end{thebibliography}
}

\end{document}